\documentclass{article}
\usepackage[preprint]{neurips_2026}
\workshoptitle{Sim2Science}
\usepackage[utf8]{inputenc}
\usepackage[T1]{fontenc}
\usepackage{hyperref}
\usepackage{url}
\usepackage{booktabs}
\usepackage{amsfonts}
\usepackage{amsmath}
\usepackage{nicefrac}
\usepackage{microtype}
\usepackage{xcolor}
\usepackage{graphicx}
\graphicspath{{figures/}}
\newcommand{\Cp}{\ensuremath{C_p}}
\newcommand{\Rsq}{\ensuremath{R^2}}
\newcommand{\ang}[1]{#1\ensuremath{^{\circ}}}
\newcommand{\panel}[1]{\textbf{(#1)}}

\newcommand{\figslot}[4]{%
  \IfFileExists{figures/#1}{%
    \includegraphics[width=#2\linewidth]{#1}%
  }{%
    \setlength{\fboxsep}{6pt}%
    \fbox{%
      \begin{minipage}[c][#3\linewidth][c]%
        {\dimexpr#2\linewidth-2\fboxsep-2\fboxrule\relax}%
        \centering\small
        \textbf{[FIGURE SLOT: \detokenize{figures/}\detokenize{#1}]}\par\medskip
        #4
      \end{minipage}}%
  }%
}
\title{A Data Fusion Framework for Grounding Aerospace Surrogate
Model via Experimental Wind-Tunnel Observations}
\author{%
  Nitin Nagesh Kulkarni\thanks{Corresponding author:
    \texttt{dheeraj@luminary.ai}} \quad
  Dheeraj Vemula \quad
  Yin Yu \quad
  Peter Lyu \quad
  Juan J. Alonso \\[4pt]
  \textbf{Luminary AI} \\
  101 S Ellsworth Ave, \\San Mateo, CA, USA \\
}
\begin{document}
\maketitle
\begin{abstract}
Aerodynamic surrogate models trained on high-fidelity Computational Fluid
Dynamics (CFD) data reproduce numerical predictions of both scalar outputs and
entire fields accurately, yet their predictive fidelity is limited by
systematic discrepancies between CFD and experimental observations. We present
an experimentally grounded correction framework that adapts a CFD-trained deep
learning surrogate using wind-tunnel Pressure-Sensitive Paint (PSP)
measurements. A Geotransolver surrogate trained on 2{,}300 high-fidelity CFD
simulations of the NASA Common Research Model (CRM) wing-body configuration,
spanning geometric variation, Mach 0.70--0.85, and angles of attack \ang{0} to
\ang{4}, reproduces the CFD integrated aerodynamic forces and pitching moment
to $\Rsq > 0.99$ but does not match the experimental data. To incorporate
experimental information without retraining the surrogate, a correction
network is trained on spatially registered PSP measurements at two freestream
Mach numbers (0.70 and 0.85) across the same angle-of-attack range, learning
the discrepancy between the surrogate-predicted and experimentally measured
surface-pressure distributions. At Mach 0.85 the correction substantially
improves agreement with PSP, particularly at the wing suction peak, shock
location, and subsequent pressure recovery, reducing both the magnitude of the
prediction error and the fraction of wetted surface on which it exceeds
$0.05$ in \Cp, and it does so from a limited experimental dataset without
modifying the pretrained surrogate parameters. On held-out angles of attack the
grounded surrogate agrees with measurement to within $2.3$--$2.7\%$ of the
measured \Cp\ range, and outperforms direct interpolation between the measured
conditions at every state tested. Experimental measurements can therefore
ground a large-scale simulation-trained surrogate by learning systematic
CFD-to-experiment discrepancies while preserving its generalization capability
and computational efficiency.
\end{abstract}
\section{Introduction}
\label{sec:intro}
High-fidelity numerical simulations, particularly Computational Fluid Dynamics
(CFD), have become indispensable for analyzing and designing complex
aerodynamic systems. However, their computational cost can make large-scale
parameter sweeps, optimization, uncertainty quantification, and real-time
prediction prohibitively expensive or infeasible. This challenge has driven
the development of data-driven surrogate models that approximate the
underlying physics at substantially lower computational cost
\cite{peherstorfer2018,giacomini2026}. Early approaches, including
Physics-Informed Neural Networks (PINNs) \cite{raissi2019,raissi2017} and Deep
Operator Networks (DeepONet) \cite{lu2019}, established the foundation for
learning physical relationships from governing equations and simulation data.
More recently, Neural Operators (NOs) and Physics-Informed Neural Operators
(PINOs) \cite{kovachki2023,li2021} have enabled the learning of mappings
between function spaces, while architectures such as Decomposable Multi-scale
Iterative Neural Operator (DoMINO) \cite{ranade2025} and Geotransolver
\cite{adams2025} have extended these capabilities to complex geometries and
large-scale problems. Despite these advances, surrogate models remain
fundamentally constrained by the fidelity of the data on which they are
trained.

The limitations of numerical simulation propagate directly to the surrogate. CFD solutions contain modeling and discretization errors from turbulence and transition models, boundary conditions, mesh resolution, and numerical approximations \cite{yeo2020,jameson1998}, particularly near shocks, suction peaks, and separation where small flow changes can strongly affect aerodynamic performance. A surrogate trained only on CFD therefore learns an approximation of the numerical model rather than the physical system, carrying simulation bias into predicted fields, quantities of interest, and design decisions. Experimental measurements can ground these predictions in physical observations: pressure taps provide localized measurements, while Pressure-Sensitive Paint (PSP) provides dense surface-pressure fields \cite{bell2011} and captures effects difficult to model accurately, including unsteadiness, manufacturing variation, facility effects, and aeroelastic deformation \cite{yasue2016,xiong2021}. However, experiments are expensive and limited across flight conditions, making them insufficient alone for training deep surrogates. This motivates combining the broad coverage of simulation with the physical fidelity of experiment \cite{giacomini2026}.
Data fusion has therefore been widely explored to combine information across fidelity levels. Classical approaches introduce additive or multiplicative correction terms \cite{kennedy2000}, while multi-fidelity surrogates use Gaussian-process co-kriging \cite{forrester2007} and composite neural networks \cite{he2020}. Other approaches reconstruct full fields from sparse measurements using proper orthogonal decomposition \cite{willcox2005,buithanh2004}, or fuse experimental and computational aerodynamic data using multi-fidelity surrogates \cite{kuya2011} and Gaussian-process models \cite{anhichem2024}. While these methods effectively reconcile information across fidelity levels, they primarily combine data through statistical or explicit correction models rather than modifying the learned representation of a pretrained surrogate.

These complementary strengths suggest that experimental data need not replace simulation, but can instead be used to refine a surrogate that already captures the broad response learned from CFD. In this work, we introduce a novel data-fusion framework that uses limited experimental observations to correct systematic discrepancies in a pretrained simulation-based surrogate. The approach preserves the surrogate's scalability, computational efficiency, and broad simulation coverage while grounding its learned representation in the physical system. This provides a closed-loop connection between simulation, surrogate modeling, and experiment, enabling experimentally grounded predictions without requiring prohibitively large experimental datasets.
\section{Method}
\label{sec:method}
\begin{figure}[t]
\centering
\begin{minipage}[c]{0.30\linewidth}
\centering
\includegraphics[width=\linewidth]{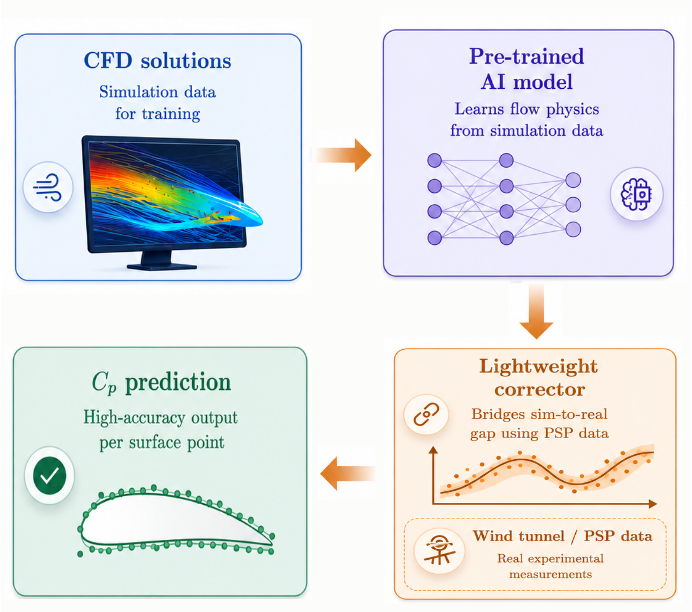}
{\small \panel{a} Grounding framework}
\end{minipage}\hfill
\begin{minipage}[c]{0.625\linewidth}
\centering
\includegraphics[width=\linewidth]{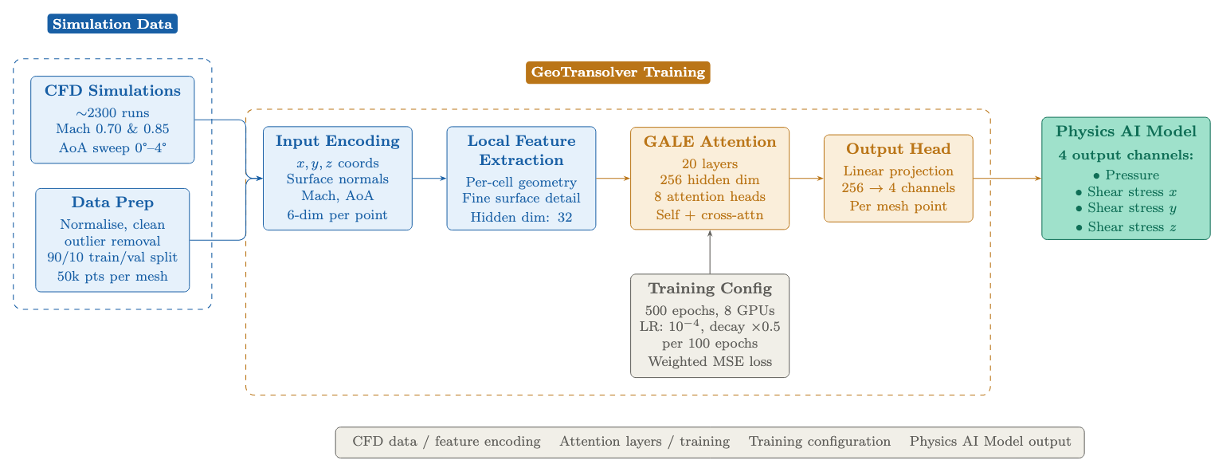}
{\small \panel{b} Baseline architecture}
\includegraphics[width=0.88\linewidth]{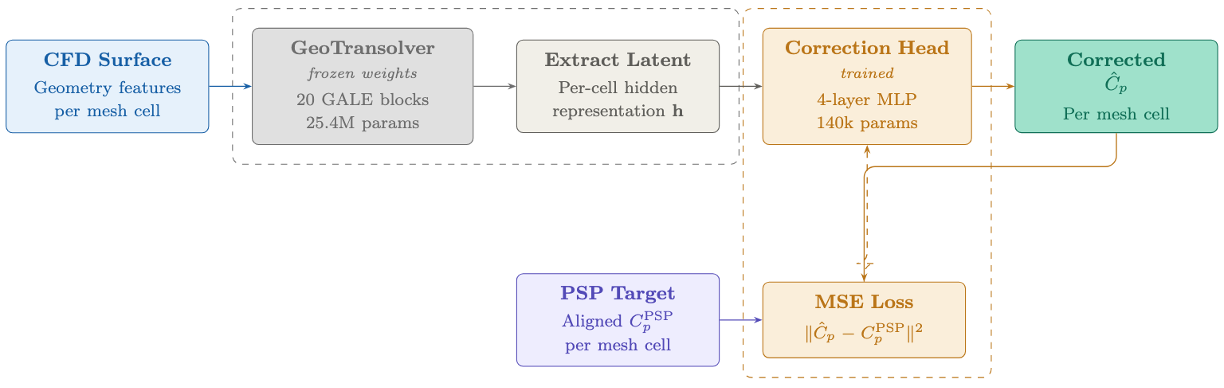}
\newline {\small \panel   {c} Correction framework}
\end{minipage}
\caption{\panel{a} Overview of the proposed simulation-to-experiment
grounding framework. \panel{b} Baseline architecture: training the surrogate
on high-fidelity CFD datasets. \panel{c} Correction framework: fine-tuning
the surrogate using experimental data to learn the CFD-to-experiment
discrepancy.}
\label{fig:framework}
\end{figure}

The proposed data-fusion methodology combines high-fidelity CFD simulations with
limited experimental measurements to correct systematic discrepancies between
simulation and experiment. The overall workflow is illustrated in
Figure~\ref{fig:framework}a. A surrogate model is first pretrained on a large
CFD dataset to learn the aerodynamic response across the prescribed
flight-condition space. The pretrained surrogate is then frozen, and a
lightweight correction head is trained using wind-tunnel PSP measurements. Rather than retraining the full surrogate, the
correction head learns the discrepancy between the CFD-based prediction and
the experimentally measured surface-pressure field.
\subsection{Datasets}
The SHIFT-Wing dataset comprises 2{,}300 high-fidelity simulations of varying NASA CRM geometry \cite{shiftwing2025}, spanning Mach numbers ($M$) of 0.70--0.85 and angles
of attack from \ang{0} to \ang{4}. These simulations provide the primary
supervision for pretraining the baseline surrogate. Experimental data are
obtained from wind-tunnel PSP measurements of the NASA CRM
\cite{bell1997}, covering Mach 0.70 and 0.85 with nine angles of attack at
each Mach number. The \ang{1.5} and \ang{3.0} conditions are withheld for
testing. The measured surface pressures are converted to pressure
coefficients, \Cp, and re-dimensionalized to the target flow conditions,
providing a consistent representation for comparison with the CFD predictions
and subsequent training of the correction head.
\subsection{Baseline Surrogate}
The baseline surrogate uses the Geotransolver \cite{adams2025} architecture, as shown in Figure~\ref{fig:framework}b. The model takes the wing surface geometry and flow conditions, including Mach number and angle of attack, as inputs and predicts the corresponding surface-pressure field. Surface mesh cells are represented using spatial coordinates, surface normals, and local geometric features. These representations are encoded and processed through geometry-aware feature extraction layers followed by GALE (Geometry-Aware Latent Embeddings) attention blocks, which capture interactions between spatially distributed surface features. The CFD data are normalized and divided into training and validation subsets. Model training and hyperparameter selection are described in Appendix~\ref{app:hyperparameters}. The resulting surrogate provides an efficient approximation of the high-fidelity CFD solution. After training, the Geotransolver parameters are frozen and its learned latent representation is used as the physics-informed prior for experimental correction.
\subsection{Experimental Correction}
During the correction stage, the frozen Geotransolver receives the corresponding surface geometry and flow conditions and produces a latent per-cell representation, $h$. These features are concatenated with the normalized flow conditions and passed to a lightweight multilayer perceptron (MLP) correction head, as shown in Figure~\ref{fig:framework}c. The correction head is trained on the spatially registered PSP measurements to predict the experimentally observed surface pressure from the frozen representation (see Appendix~\ref{app:head}). Thus, the trainable component is restricted to the correction head, while the pretrained Geotransolver remains fixed. This formulation allows the extensive CFD dataset to provide the underlying aerodynamic representation while using experimental observations at a small number of conditions to learn systematic CFD-to-experiment discrepancies. The correction-head formulation and PSP-to-CFD surface registration procedure are provided in Appendices~\ref{app:head} and \ref{app:registration}, respectively.
\section{Results}
\label{sec:results}
\begin{figure}[t] \centering \includegraphics[width=0.8\linewidth]{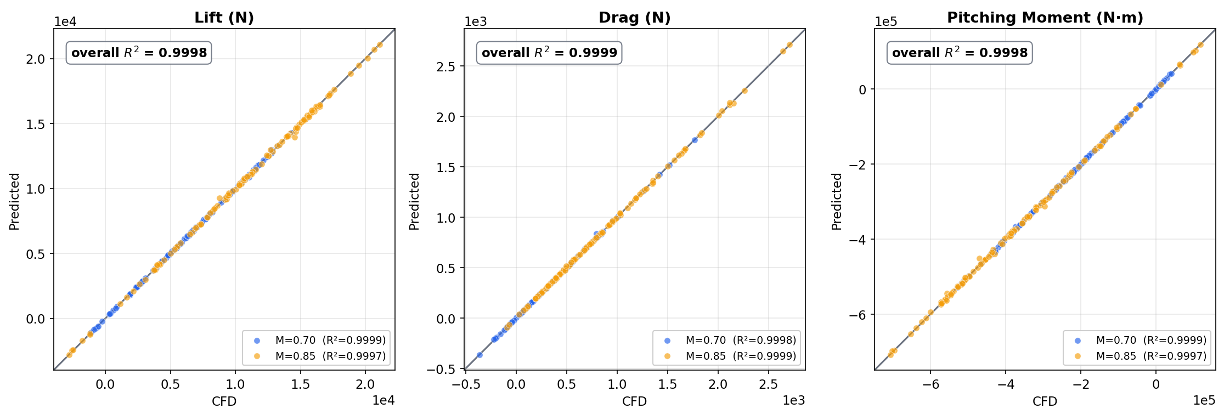} \caption{Lift, drag and pitching moments: baseline geotransolver surrogate model performance across the Mach 0.70 to 0.85 and AoA \ang{0} to \ang{4} flight envelope.} \label{fig:loads} \end{figure}

\begin{figure}[b] \centering \begin{minipage}[t]{0.47\linewidth} \centering \includegraphics[width=\linewidth]{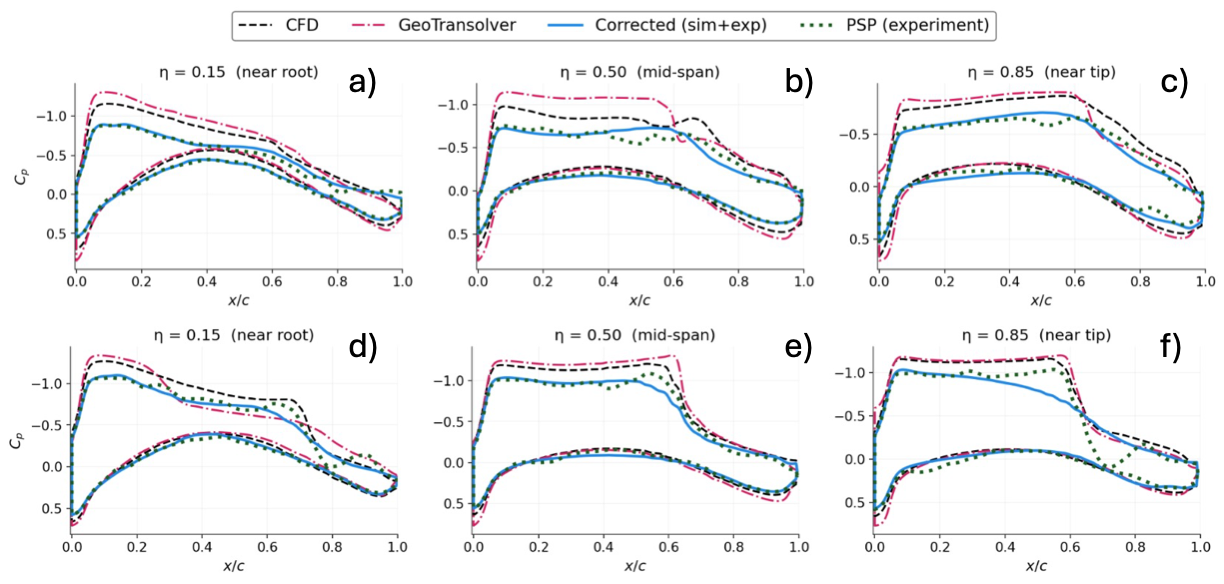}\\[3pt] {\small \panel{a} Spanwise \Cp\ distribution} \end{minipage}\hfill \begin{minipage}[t]{0.52\linewidth} \centering \includegraphics[width=\linewidth]{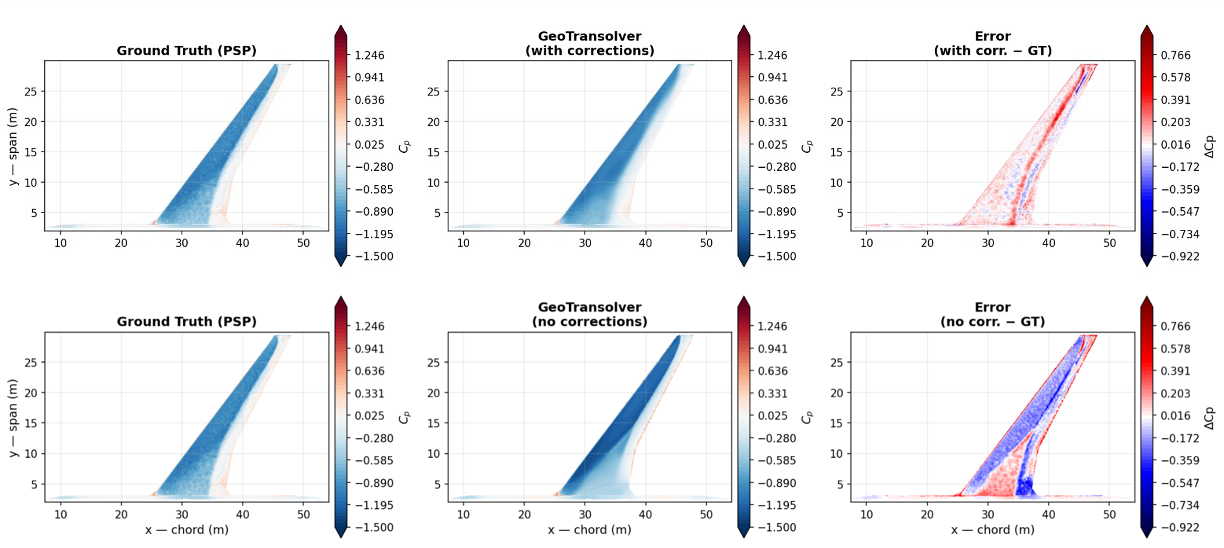}\\[3pt] {\small \panel{b} Full-surface \Cp\ fields and errors} \end{minipage} \caption{\panel{a} Spanwise \Cp\ distributions at $M=0.85$ comparing CFD, PSP, and corrected predictions at AoA \ang{1.5} (a--c) and \ang{3.0} (d--f) from near-root to near-tip. \panel{b} Surface \Cp\ fields from PSP, uncorrected, and corrected predictions.}
 \label{fig:results085} \end{figure}

The first objective is to establish whether the surrogate faithfully reproduces
the high-fidelity CFD solution used for training. Across the transonic flight
envelope (Mach 0.70--0.85 and angles of attack from \ang{0} to \ang{4}),
Figure~\ref{fig:loads} compares the predicted and CFD-based integrated
aerodynamic loads. The Geotransolver achieves $\Rsq > 0.99$ for lift, drag,
and pitching moment, demonstrating that it accurately captures the global
aerodynamic response encoded in the CFD data with negligible additional error.
This establishes the CFD-trained surrogate as a controlled baseline for
evaluating the subsequent experimental correction. Importantly, agreement with
CFD does not constitute experimental validation; because the model is trained
exclusively on CFD, any systematic discrepancy between CFD and the physical
system is inherited by the surrogate.
Although the baseline closely reproduces CFD, comparison with PSP reveals
localized discrepancies in the surface-pressure distribution. At $M=0.85$,
Figure~\ref{fig:results085}a compares the pressure coefficient (\Cp) at root,
mid-span, and near-tip stations. The largest differences occur near the
suction peak and downstream pressure recovery, where the CFD-predicted shock
location and pressure rise increasingly diverge from the experimental
measurements toward the mid- and outboard-span regions. The experimental
correction shifts the predicted \Cp\ distributions toward the PSP measurements
while preserving the primary aerodynamic features learned from CFD. This
demonstrates that sparse experimental data can capture the systematic
discrepancy between the numerical and physical systems without discarding the
underlying CFD-derived flow representation. Results at Mach 0.70 are provided
in Appendix~\ref{app:results}.
\subsection{Quantitative agreement and interpolation baseline}
\label{sec:metrics}
Table~\ref{tab:metrics} summarizes agreement between the grounded surrogate and
registered PSP fields at the four held-out conditions. Because \Cp\ is signed
and passes through zero, pointwise relative errors such as mean absolute
percentage error are not well defined. We therefore normalize MAE by the
measured \Cp\ range at each condition. The grounded surrogate achieves
2.3--2.7\% normalized error across all conditions, with $\Rsq=0.922$--$0.966$.
\begin{table}[h]
\centering
\caption{Per-condition agreement between the grounded surrogate and PSP at held-out states. Normalized error is MAE divided by the measured \Cp\ range.
}
\label{tab:metrics} \small \begin{tabular}{llccccc} \toprule $M$ & AoA & \Rsq & MAE & RMSE & Norm.\ err.\ (\%) & \Cp\ range \\ \midrule 0.70 & \ang{1.5} & 0.940 & 0.059 & 0.090 & 2.4 & 2.44 \\ 0.70 & \ang{3.0} & 0.966 & 0.065 & 0.100 & 2.3 & 2.82 \\ 0.85 & \ang{1.5} & 0.922 & 0.069 & 0.110 & 2.5 & 2.75 \\ 0.85 & \ang{3.0} & 0.952 & 0.070 & 0.110 & 2.7 & 2.64 \\ \bottomrule \end{tabular} \end{table}
Because the held-out angles of attack are bracketed by measured conditions, we
next test whether the correction provides information beyond direct
interpolation of neighboring PSP fields. Two interpolation baselines are
considered: a \emph{raw} baseline using the PSP data on its native structured
grid without coordinate registration, and a \emph{registered} baseline using
the fields produced by the registration procedure in
Appendix~\ref{app:registration}. This comparison separates the contribution
of coordinate registration from that of the learned correction.
As shown in Table~\ref{tab:interp}, registration alone improves $\Rsq$ by
$0.10$--$0.16$, confirming that surface registration is a substantive step.
The correction head provides a further $0.09$--$0.20$ improvement over the
registered baseline, with the grounded surrogate outperforming interpolation
at every held-out condition. The distinction is most pronounced at
$M=0.85$, AoA $=\ang{3.0}$, where interpolation falls to $\Rsq=0.756$ while
the grounded surrogate achieves $\Rsq=0.952$, a margin of $0.196$. Thus, the
improvement cannot be attributed to registration or smooth interpolation in
$(M,\alpha)$ alone, and is largest in the strongly transonic regime where the
shock position varies most rapidly.
\begin{table}[h]
\centering
\caption{Surface \Cp\ agreement with PSP at held-out conditions. Interpolation uses raw, registered PSP fields, while the grounded surrogate combines the frozen backbone with the correction head.}
\label{tab:interp} \small \begin{tabular}{lccccc} \toprule & \multicolumn{2}{c}{Interpolation baseline} & Grounded & \multicolumn{2}{c}{$\Delta \Rsq$ from} \\ \cmidrule(lr){2-3} \cmidrule(lr){5-6} Condition & raw & registered & surrogate & registration & correction \\ \midrule $M = 0.70$, AoA $=$ \ang{1.5} & 0.743 & 0.849 & \textbf{0.940} & $+0.106$ & $+0.091$ \\ $M = 0.70$, AoA $=$ \ang{3.0} & 0.710 & 0.874 & \textbf{0.966} & $+0.164$ & $+0.092$ \\ $M = 0.85$, AoA $=$ \ang{1.5} & 0.722 & 0.828 & \textbf{0.922} & $+0.106$ & $+0.094$ \\ $M = 0.85$, AoA $=$ \ang{3.0} & 0.655 & 0.756 & \textbf{0.952} & $+0.101$ & $+0.196$ \\ \bottomrule \end{tabular} \end{table}
The observed improvements indicate that the correction captures physical discrepancies absent from the rigid-geometry CFD training data, particularly in shock location, suction-peak magnitude, and pressure recovery. One possible contributor is aeroelastic deformation, as the physical wind-tunnel model can deform under aerodynamic loading, altering local incidence and surface geometry; this effect is quantified in Appendix~\ref{app:aeroelastic}. Other contributors may include transition behavior, turbulence modeling, facility conditions, and model-form uncertainties. The correction does not isolate these sources but learns their combined effect from experimental observations. Thus, rather than replacing CFD or explicitly modeling each physical mechanism, the method grounds the simulation-trained representation in the physical system while retaining its broad aerodynamic structure and computational efficiency. The scope and remaining limitations are discussed in Appendix~\ref{app:limitations}.
\section{Conclusion}
This study introduced an experimental grounding framework for simulation-trained
surrogates that combines a lightweight correction model with a pretrained
Geotransolver. Across Mach 0.70--0.85 and angles of attack from \ang{0} to
\ang{4}, the baseline surrogate reproduced CFD-derived integrated aerodynamic
loads with $\Rsq > 0.999$. However, comparison with wind-tunnel PSP revealed
localized discrepancies in surface pressure, particularly near suction peaks,
pressure-recovery regions, and transonic shocks. By freezing the pretrained
backbone and training a compact correction head on registered PSP
measurements, the proposed approach reduced these discrepancies without
retraining the full surrogate. At the held-out conditions, the corrected
predictions achieved errors of 2.3--2.7\% relative to the measured \Cp\ range
and outperformed direct interpolation between measured conditions at every
tested state. These results demonstrate that limited experimental data can
effectively ground a simulation-trained surrogate, reducing systematic
CFD-to-experiment discrepancies while retaining the broad aerodynamic
representation and computational efficiency of the original model. Determining
the extent to which the correction transfers beyond the measured flow envelope
remains an important direction for future work.
\newpage
\bibliographystyle{unsrtnat}
\bibliography{references}
\newpage

\appendix
\newpage
\section{Appendix}
\subsection{Training dataset and hyperparameters}
\label{app:hyperparameters}
The training SHIFT-Wing dataset consists of high-fidelity
RANS solutions of the NASA CRM wing-body configuration, 1{,}130 at Mach 0.70
and 1{,}132 at Mach 0.85, drawn from 2{,}330 generated cases.
The dataset is not a sweep of flow conditions over a single fixed geometry: it
varies the wing geometry parametrically alongside the flow conditions, so that
the surrogate learns the aerodynamic response as a function of shape as well as
flight condition. Geometry is defined by a parametric CAD model and sampled by
Latin hypercube over the seven design variables listed in
Table~\ref{tab:doe}, with angle of attack swept over the same range and
sideslip held at zero.
\begin{table}[h]
  \centering
  \caption{Design of experiments for the SHIFT-Wing dataset. The seven
  geometry variables are sampled by Latin hypercube; angle of attack is swept
  and sideslip is fixed.}
  \label{tab:doe}
  \small
  \begin{tabular}{llc}
    \toprule
    Variable & Meaning & Range \\
    \midrule
    $AR$                     & Aspect ratio                  & 7.5 -- 11 \\
    $\Lambda_{c/4}$          & Quarter-chord sweep (\ang{})  & 25 -- 37.5 \\
    $f_{c_r}$                & Root-chord extension factor   & 1.0 -- 1.4 \\
    $D_f$                    & Fuselage diameter (m)         & 6.096 -- 6.553 \\
    $\tau_r$                 & Root twist (\ang{})           & 3 -- 9 \\
    $\Delta\tau_k$           & Kink twist increment (\ang{}) & $-7$ to $-3$ \\
    $\Delta\tau_t$           & Tip twist increment (\ang{})  & $-7.5$ to $-1.5$ \\
    \midrule
    $\alpha$                 & Angle of attack (\ang{})      & 0 -- 4 \\
    $\beta$                  & Sideslip (\ang{})             & 0 (fixed) \\
    \bottomrule
  \end{tabular}
\end{table}
Solutions were computed with the Luminary Cloud solver as compressible steady
RANS, using an ideal-gas model (molecular weight 28.96, $c_p =
1004.703$~J\,kg$^{-1}$K$^{-1}$), Sutherland viscosity, $\mathrm{Pr} = 0.72$,
and a high-accuracy spatial scheme. Turbulence is closed with the
Spalart-Allmaras one-equation model initialized from freestream. Meshes are
adaptively refined all-tetrahedral grids with a target of approximately
$30 \times 10^6$ cells, an initial boundary-layer cell size of
$5 \times 10^{-6}$ and a growth ratio of 1.2. Cases were run to a residual
threshold of $10^{-8}$ with a maximum of 15{,}000 iterations. Freestream
conditions correspond to 35{,}000~ft ($T_\infty = 218.81$~K, $p_\infty =
23{,}840$~Pa, $\rho_\infty \approx 0.380$~kg\,m$^{-3}$, $\mu_\infty \approx
1.43 \times 10^{-5}$~Pa\,s), with velocity set by the Mach number. Because the
altitude is fixed and the geometry varies, Reynolds number is not held
constant across the dataset: it varies with mean aerodynamic chord and is of
order $10^{7}$. Preprocessing included outlier removal, normalization, and a
90/10 train/validation split.
The baseline was trained on 8 NVIDIA H100 GPUs for 500 epochs with an initial
learning rate of $10^{-4}$, a learning rate decay factor of 0.5 applied every
100 epochs, and a mean-squared-error loss on the normalized surface pressure
coefficient, weighted per cell by cell area so that the loss is not dominated
by regions of fine mesh refinement. The correction head is the four-layer perceptron
$386 \rightarrow 256 \rightarrow 128 \rightarrow 64 \rightarrow 1$ with SiLU
activations, giving 140{,}289 trainable parameters. Its 386 inputs are the
384-dimensional per-cell latent from the final Geotransolver block
concatenated with Mach number and angle of attack, and its single output is the
normalized pressure coefficient. It was trained for 200 epochs with AdamW
(learning rate $3 \times 10^{-4}$, weight decay $10^{-5}$), a cosine-annealing
schedule over the full run, and batches of 16{,}384 per-cell points pooled
across conditions, minimizing the unweighted mean squared error of
Eq.~\eqref{eq:loss}. The backbone is frozen throughout and consumes
pre-extracted latents. Of the 18 experimental conditions, 14 were used for
training, the two \ang{1.5} conditions for validation, and the two \ang{3.0}
conditions held out for testing.
\subsection{Formulation of the correction head}
\label{app:head}
For surface cell $i$ with geometric descriptor $x_i$ and flow condition
$m = (M, \alpha)$, the frozen backbone with parameters $\theta$ produces a
per-cell latent representation and the baseline decoder reproduces the CFD
field,
\begin{equation}
  h_i = \mathcal{G}_\theta(x_i, m),
  \qquad
  \hat{C}_{p,i}^{\mathrm{sim}} = \mathcal{D}_\theta(h_i).
  \label{eq:backbone}
\end{equation}
The correction head $g_\phi$, a compact multilayer perceptron, takes the
384-dimensional latent concatenated with the normalized flow condition
$\bar{m} = (M, \alpha)$ and predicts the experimentally observed pressure,
\begin{equation}
  \hat{C}_{p,i}^{\mathrm{exp}} = g_\phi\!\left(\left[h_i,\, \bar{m}\right]\right),
  \label{eq:head}
\end{equation}
fitted by minimizing the mean squared error against the registered
measurements over the set $\mathcal{S}$ of valid wing-surface cells,
\begin{equation}
  \mathcal{L}(\phi) = \frac{1}{|\mathcal{S}|}
  \sum_{i \in \mathcal{S}}
  \left( \hat{C}_{p,i}^{\mathrm{exp}} - C_{p,i}^{\mathrm{PSP}} \right)^2,
  \qquad \theta \ \text{fixed}.
  \label{eq:loss}
\end{equation}
The loss is unweighted and acts on surface \Cp\ alone; no CFD anchoring term is
used, since the CFD information enters entirely through the frozen backbone.
Equation~\eqref{eq:head} is not restricted to the additive form
$\hat{C}_p^{\mathrm{sim}} + \delta_\phi$ used in classical model-correction
formulations, in which a low-fidelity model is corrected toward a trusted one
through prescribed additive and multiplicative terms \cite{kennedy2000}. The
head implemented here instead learns an unconstrained nonlinear mapping from
the frozen latent representation to the measured pressure, which can represent
additive and multiplicative corrections as special cases without assuming
either. What limits the experimental data requirement is therefore not a
restricted correction form but the fact that $\theta$ remains fixed: the
aerodynamic representation is inherited from the full CFD dataset, and only the
140{,}289 parameters of $g_\phi$ are fitted to experiment.
\subsection{PSP registration procedure}
\label{app:registration}
\begin{figure}[h]
  \centering
  \figslot{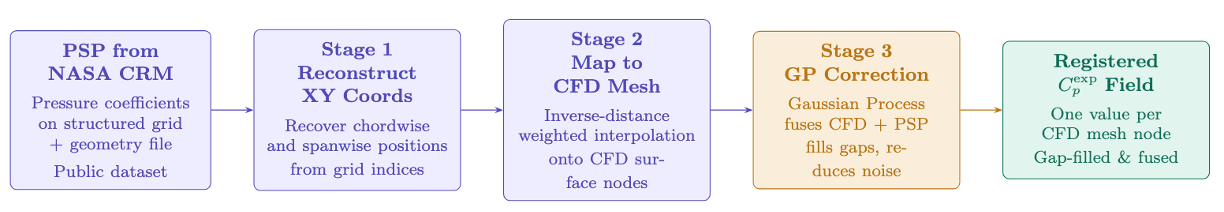}{0.95}{0.26}{Drop the joined registration figure into the figures directory under this name. Expected content: PSP input, Stage 1 coordinate reconstruction, Stage 2 mapping to the CFD mesh, Stage 3 GP correction, and the registered Cp field. This box disappears automatically once the file is present.}
  \caption{Coordinate registration pipeline. The delivered PSP field is indexed
  by structured grid position, so chordwise and spanwise coordinates are first
  reconstructed from the grid indices (Stage~1); the resulting measurement
  cloud is transferred onto the CFD surface nodes by inverse-distance weighted
  interpolation (Stage~2); and a Gaussian process is fitted over the surface to
  suppress measurement noise and fill nodes left without a valid paint return
  (Stage~3). The output is a registered $C_p^{\mathrm{exp}}$ field carrying one
  value per CFD mesh cell, which is the target used for fitting and evaluating
  the correction head. Cells with no valid PSP return are masked and excluded
  from both.}
  \label{fig:registration}
\end{figure}

The PSP measurements and the CFD solutions are defined on different geometric
domains, and the experimental dataset provides no fixed global coordinate
system, so the two must be brought onto a common surface before any pointwise
comparison is meaningful. The measured field is projected, registered, and
mapped onto the CFD surface mesh, yielding one experimental pressure value per
mesh cell that can be used directly as a training target. The pipeline is shown
in Figure~\ref{fig:registration}. It takes as input the public NASA CRM PSP
dataset, which supplies pressure coefficients on a structured grid together
with a geometry file, and proceeds in three stages.

\textbf{Stage 1: reconstruct surface coordinates.} The delivered PSP data are
indexed by structured grid position rather than by physical location, so
chordwise and spanwise positions are first recovered from the grid indices
using the accompanying geometry file, placing every measurement at a physical
point on the wing surface. Until this step is performed the measurement carries
no usable spatial reference, and the resulting alignment error is large: an
interpolation baseline built on the unregistered field loses $0.10$--$0.16$ in
\Rsq\ relative to the same baseline built on the registered field
(Table~\ref{tab:interp}, raw versus registered columns).
\textbf{Stage 2: map to the CFD mesh.} The reconstructed measurement cloud is
transferred onto the CFD surface nodes by inverse-distance weighted
interpolation, giving a first estimate of $C_p^{\mathrm{exp}}$ at every mesh
node within the painted region. The measurement grid and the CFD surface mesh
are unrelated discretizations, so this step resamples rather than reindexes,
and its accuracy is limited where the two differ most in local resolution.
\textbf{Stage 3: Gaussian process correction.} A Gaussian process is then fitted
over the surface to reduce measurement noise and to fill nodes left without a
valid PSP return. The result is a registered $C_p^{\mathrm{exp}}$ field carrying one value per CFD
mesh cell. Two properties of the experimental data constrain this step. First,
PSP coverage of the surface is incomplete: optical access limits leave gaps at
edges and at component junctions, so cells without a valid paint response are
masked and excluded from both fitting and evaluation. Second, the as-built
model geometry and its aeroelastic deflection at test conditions both deviate
from the nominal CAD shape on which the CFD mesh is built, so the registration
aligns surfaces that are not exactly congruent;
Appendix~\ref{app:aeroelastic} quantifies the magnitude of this deviation.
Registration is therefore not a neutral preprocessing step: it carries its own
error, and the raw-versus-registered comparison in Table~\ref{tab:interp}
bounds how much of the reported agreement depends on it.
\subsection{Aeroelastic wing deflection}
\label{app:aeroelastic}
The CFD simulations are performed on the nominal CAD wing geometry, whereas
the PSP measurements are acquired on a flexible model that deforms under
aerodynamic load, bending upward and twisting nose-down (washout) relative to
its unloaded shape. The magnitude of the deformation scales approximately with
the product of freestream dynamic pressure and lift coefficient,
$q_\infty C_L$. Published measurements on the NASA CRM report tip bending of
approximately 25~mm and tip twist of approximately \ang{-0.5} at the
$M = 0.85$ cruise design point \cite{yasue2016,keye2015,xiong2021}.
A portion of the observed CFD-to-PSP discrepancy is therefore attributable to
a geometric mismatch between the two pipelines rather than to turbulence or
transition modeling deficiencies. Washout unloads the mid-span and tip
regions, bringing the upper- and lower-surface \Cp\ distributions closer
together, which reduces the suction-peak magnitude and lift, and shifts the
shock forward toward the leading edge or removes it altogether. This is
consistent with the spanwise pattern in Figure~\ref{fig:results085}a, where the
discrepancy between the uncorrected prediction and the measurements grows from
root to tip, following the outboard growth of the twist.
\subsection{Additional results}
\label{app:results}
\begin{figure}[h]
  \centering
  \includegraphics[width=\linewidth]{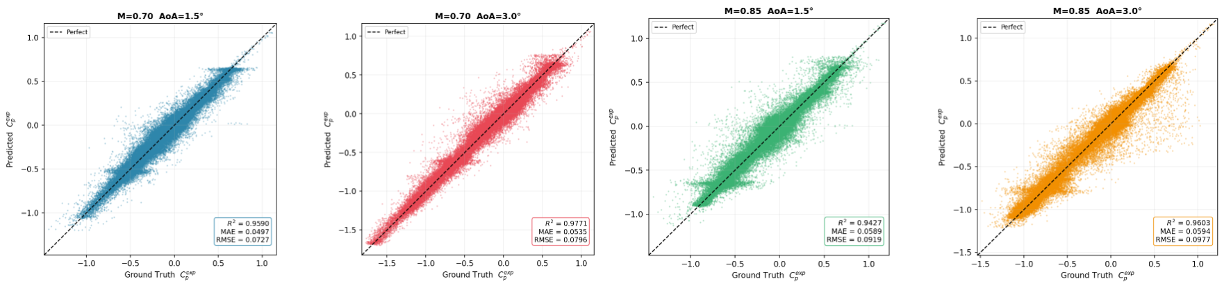}
  \caption{Surface \Cp\ parity plots: predicted versus experimental ground
  truth \Cp\ across Mach 0.70 and 0.85 at AoA $=$ \ang{1.5} and \ang{3.0}.
  }
  \label{fig:parity}
\end{figure}

\begin{figure}[h]
  \centering
  \includegraphics[width=0.85\linewidth]{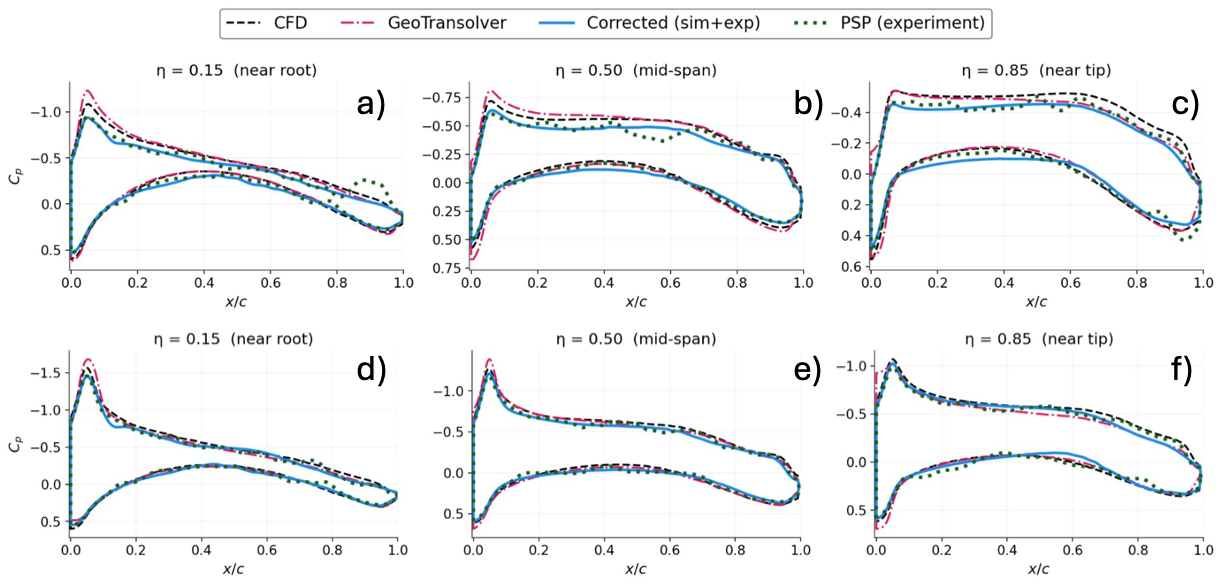}
  \caption{Spanwise \Cp\ distribution: comparison of baseline CFD,
  experimental PSP measurements, and calibrated surrogate model predictions at
  Mach 0.70 across spanwise sections ($\eta = 0.15$ near root, $\eta = 0.50$
  mid-span, and $\eta = 0.85$ near tip) for AoA $=$ \ang{1.5} (a--c) and AoA
  $=$ \ang{3.0} (d--f).}
  \label{fig:spanwise070}
\end{figure}

\begin{figure}[h]
  \centering
  \begin{minipage}[t]{0.485\linewidth}
    \centering
    \includegraphics[width=\linewidth]{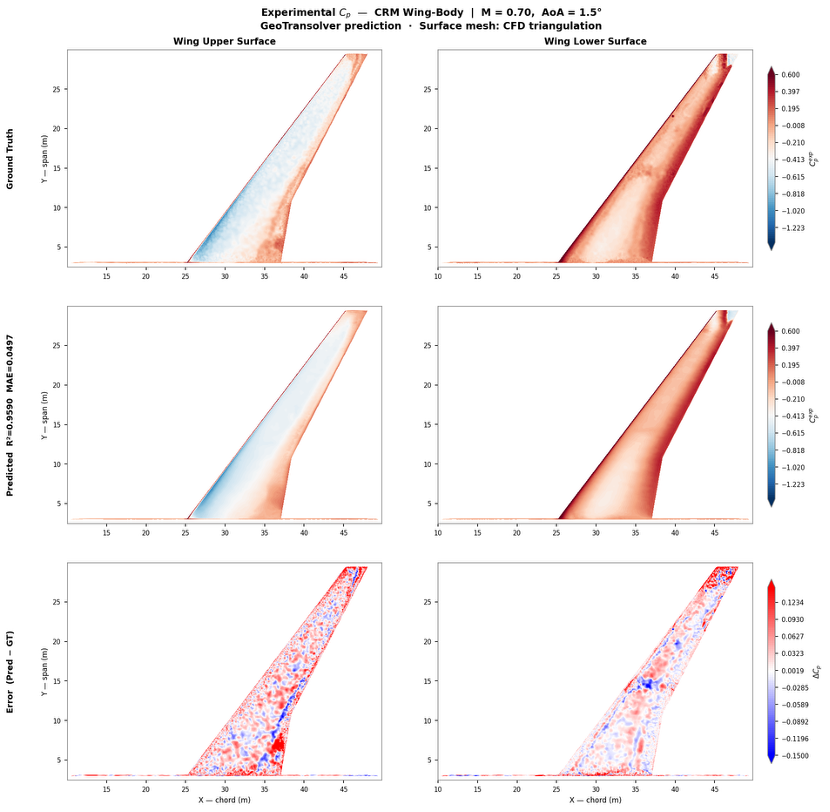}\\[3pt]
    {\small \panel{a} AoA $=$ \ang{1.5}}
  \end{minipage}\hfill
  \begin{minipage}[t]{0.485\linewidth}
    \centering
    \includegraphics[width=\linewidth]{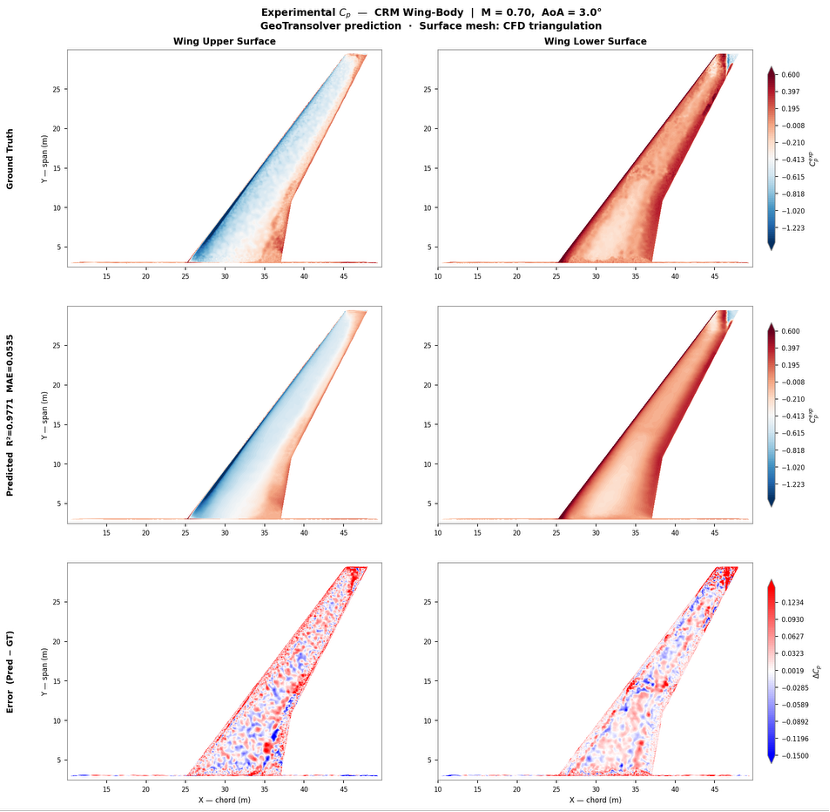}\\[3pt]
    {\small \panel{b} AoA $=$ \ang{3.0}}
  \end{minipage}
  \caption{Full-surface \Cp\ contours and error maps at Mach 0.70: comparison
  of experimental ground truth, Geotransolver prediction, and spatial error
  field. \panel{a} AoA $=$ \ang{1.5} ($\Rsq = 0.940$, MAE $= 0.059$).
  \panel{b} AoA $=$ \ang{3.0} ($\Rsq = 0.966$, MAE $= 0.065$).}
  \label{fig:contours070}
\end{figure}

Evaluation metrics for the calibrated surrogate model at Mach 0.70 across
representative angles of attack are as follows. Figure~\ref{fig:parity} shows parity scatter plots comparing predicted surface
pressure coefficients with wind-tunnel experimental ground truth across four
flow conditions ($M = 0.70$ and $M = 0.85$ at AoA $=$ \ang{1.5} and
\ang{3.0}). Across all evaluated conditions the calibrated surrogate
predictions lie close to the 1:1 line, with area-weighted
$\Rsq = 0.938$--$0.953$, MAE $= 0.046$--$0.061$ and RMSE
$= 0.063$--$0.089$ in \Cp. Evaluated per cell without area weighting, the
same predictions give the values in Table~\ref{tab:metrics}. This shows
consistent accuracy across both measured Mach numbers and both held-out angles
of attack. We note that consistency at these conditions is evidence of accuracy
within the measured envelope, not of generalization beyond it; see
Appendix~\ref{app:limitations}.
The spanwise pressure coefficient distributions at Mach 0.70 for AoA $=$
\ang{1.5} and \ang{3.0} are shown in Figure~\ref{fig:spanwise070} across the
root ($\eta = 0.15$), mid-span ($\eta = 0.50$), and tip ($\eta = 0.85$)
sections, where $\eta = y/(b/2)$ denotes the fraction of semi-span. While the
baseline CFD model exhibits minor deviations from experimental PSP data in
capturing suction peaks and trailing-edge pressure recovery, the calibrated
surrogate model corrects these localized discrepancies. The corrected pressure
distributions align with the wind-tunnel PSP measurements across all three
spanwise stations, confirming that the latent correction head captures the
physical flow features under subsonic and mild transonic conditions.
Full-surface pressure coefficient contour maps and absolute error
distributions at Mach 0.70 are shown in Figure~\ref{fig:contours070} for
AoA $=$ \ang{1.5} and \ang{3.0}. A comparison between the ground truth PSP
data and the calibrated Geotransolver predictions demonstrates strong spatial
agreement across both upper and lower wing surfaces. The error contour maps
(Pred $-$ GT) show minimal residuals across most of the wing planform, with
minor localized differences concentrated near the leading edge and wing tip.
These full-surface comparisons confirm that the fine-tuning correction
framework preserves global pressure field topology while resolving systematic
CFD-to-experiment discrepancies.
\subsection{Limitations}
\label{app:limitations}
Primary limitations of the claims made above are as follow. First, the correction is fitted
and evaluated at two Mach numbers only, so its behaviour outside
$0.70 \le M \le 0.85$ is untested; the results demonstrate accuracy at
held-out angles of attack within the measured envelope, not extrapolation
beyond it. Second, PSP data are available for a single configuration, whereas the
backbone is trained across parametric geometry variation, so whether a
correction learned on one geometry transfers to others in the SHIFT-Wing
family remains open. Third, the correction is empirical: it learns the combined
effect of all sources of CFD-to-experiment discrepancy as observed in the
measurements, and does not identify or separate individual mechanisms, so the
aeroelastic contribution of Appendix~\ref{app:aeroelastic} is offered as a
physical interpretation rather than an identified term. Fourth, results are
reported as ranges across the four evaluated flow conditions rather than as
variance over training seeds. We regard the first two as the primary questions
for extending the framework.
\end{document}